\documentclass[runningheads]{llncs}
\usepackage{epsfig}
\usepackage{graphicx}
\usepackage{tikz}
\usepackage{comment}
\usepackage{amsmath,amssymb} 
\usepackage{color}
\usepackage{caption}

\usepackage{xspace}
\makeatletter
\DeclareRobustCommand\onedot{\futurelet\@let@token\@onedot}
\def\@onedot{\ifx\@let@token.\else.\null\fi\xspace}

\def\eg{\emph{e.g}\onedot} 
\def\ie{\emph{i.e}\onedot}

\def\etal{\emph{et al}\onedot}
\makeatother

\begin{document}
\pagestyle{headings}
\mainmatter
\def\ECCVSubNumber{1291}  

\title{Open-Edit: Open-Domain Image Manipulation with Open-Vocabulary Instructions} 

\titlerunning{Open-Domain Image Manipulation with Open-Vocabulary Instructions}
%
\author{Xihui Liu\inst{1}\thanks{This work was done during Xihui Liu's internship at Adobe.} \and
Zhe Lin\inst{2} \and
Jianming Zhang\inst{2} \and
Handong Zhao\inst{2} \and
Quan Tran\inst{2} \and
Xiaogang Wang\inst{1} \and
Hongsheng Li\inst{1}}
\authorrunning{X. Liu et al.}
%
\institute{The Chinese University of Hong Kong
\email{\{xihuiliu,xgwang,hsli\}@ee.cuhk.edu.hk}\\
 \and
Adobe Research
\email{\{zlin,jianmzha,hazhao,qtran\}@adobe.com}}

\makeatletter
\let\@oldmaketitle\@maketitle
\renewcommand{\@maketitle}{\@oldmaketitle
\centering
\includegraphics[width=1\linewidth]{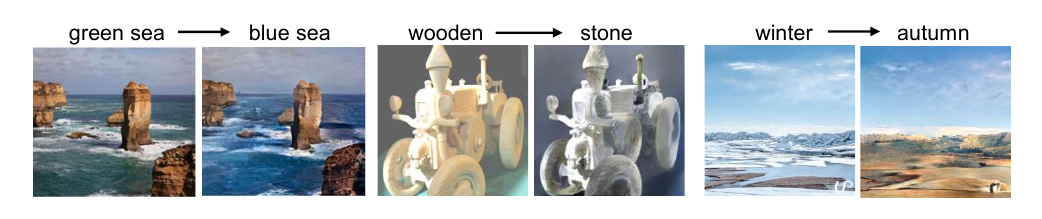}
\captionof{figure}{Examples of open-edit. The editing instruction (top), source image (left), and manipulated image (right) are shown for each example. Our approach edits open-vocabulary color, texture, and semantic attributes of open-domain images.}
\label{fig:intro}
\bigskip\bigskip}
\makeatother

\maketitle

\begin{abstract}
  We propose a novel algorithm, named Open-Edit, which is the first attempt on open-domain image manipulation with open-vocabulary instructions. It is a challenging task considering the large variation of image domains and the lack of training supervision. Our approach takes advantage of the unified visual-semantic embedding space pretrained on a general image-caption dataset, and manipulates the embedded visual features by applying text-guided vector arithmetic on the image feature maps. A structure-preserving image decoder then generates the manipulated images from the manipulated feature maps. We further propose an on-the-fly sample-specific optimization approach with cycle-consistency constraints to regularize the manipulated images and force them to preserve details of the source images. Our approach shows promising results in manipulating open-vocabulary color, texture, and high-level attributes for various scenarios of open-domain images.\footnote{Code is released at https://github.com/xh-liu/Open-Edit.}
\end{abstract}
\section{Introduction}

Automatic image editing, aiming at manipulating images based on the user instructions, is a challenging problem with extensive applications. 
It helps users to edit photographs and create art works with higher efficiency. 

Several directions have been explored towards image editing with generative models. 
Image-to-image translation~\cite{isola2017image,zhu2017unpaired,choi2018stargan} translates an image from a source domain to a target domain.
But it is restricted to the predefined domains, and cannot be generalized to manipulating images with arbitrary instructions.
%
%
%
%
GAN Dissection~\cite{bau2018gan} and GANPaint~\cite{bau2019semantic} are able to add or remove certain objects by manipulating related units in the latent space.
However, they are limited to editing a small number of pre-defined objects and stuff that can be identified by semantic segmentation and can be disentangled in the latent space.
%

%
Most relevant to our problem setting is language-based image editing~\cite{zhu2017your,dong2017semantic,nam2018text,gunel2018language,mao2019bilinear}.
%
Some previous work~\cite{el2018keep,cheng2018sequential,chen2018language} annotates the manipulation instructions and ground-truth manipulated images for limited images and scenarios. But it is infeasible to obtain such annotations for large-scale datasets.
To avoid using ground-truth manipulated images, other work~\cite{zhu2017your,dong2017semantic,nam2018text,gunel2018language,mao2019bilinear} only use images and caption annotations as training data.
Given an image A and a mismatched caption B, the model is required to edit A to match B.
The manipulated images are encouraged to be realistic and to match the manipulation instructions, without requiring ground-truth manipulated images as training supervision.
However, it is assumed that any randomly sampled caption is a feasible manipulation instruction for the image.
For example, given an image of a red flower, we can use ``a yellow flower'' as the manipulation instruction. But it is meaningless to use ``a blue bird'' as the manipulation instruction for the image of a red flower.
So this approach is restricted to datasets from a specific domain (\eg, flowers or birds in previous work~\cite{mo2018instagan}) with human-annotated fine-grained descriptions for each image, and cannot generalize to open-domain images.

In this work, we aim to manipulate open-domain images by open-vocabulary instructions with minimal supervision, which is a challenging task and has not been explored in previous work.
We propose \textit{Open-Edit}, which manipulates the visual feature maps of source images based on the open-vocabulary instructions, and generates the manipulated images from the manipulated visual feature maps.
It takes advantages of the universal visual-semantic embedding pretrained on a large-scale image-caption dataset, Conceptual Captions~\cite{sharma2018conceptual}.
The visual-semantic embedding model encodes any open-domain images and open-vocabulary instructions into a joint embedding space.
Features within the joint embedding space can be used for localizing instruction-related regions of the input images and for manipulating the related visual features.
The manipulations are performed by vector arithmetic operations between the visual feature maps and the textual features, \eg, visual embedding of green apple = visual embedding of red apple - textual embedding of ``\textit{red apple}'' + textual embedding of ``\textit{green apple}''.
Then a structure-preserving image decoder generates the manipulated images based on the manipulated visual feature maps.
The image generator is trained with image reconstruction supervision and does not require any paired manipulation instruction for training.
So our approach naturally handles open-vocabulary open-domain image manipulations with minimal supervision.

Moreover, to better preserve details and regularize the manipulated images, we introduce \textit{sample-specific optimization} to optimize the image decoder with the specific input image and manipulation instruction.
Since we cannot apply direct supervisions on the manipulated images, we adopt reconstruction and cycle-consistency constraints to optimize the small perturbations added to the intermediate decoder layers.
The reconstruction constraint forces the image decoder to reconstruct the source images from their visual feature maps;
The cycle-consistency constraint performs a cycle manipulation (\eg, red apple $\rightarrow$ green apple $\rightarrow$ red apple) and forces the final image to be similar to the original ones.

Our proposed framework, Open-Edit, is the first attempt for open-domain image manipulation with open-vocabulary instructions, with several unique advantages: 
(1) Unlike previous approaches that require single-domain images and fine-grained human-annotated descriptions, we only use noisy image-captions pairs harvested from the web for training. Results in Fig.~\ref{fig:intro} demonstrates that our model is able to manipulate open-vocabulary colors, textures, and semantic attributes of open-domain images. 
(2) By controlling the coefficients of the vector arithmetic operation, we can smoothly control the manipulation strength and achieve visual appearances with interpolated attributes.
(3) The sample-specific optimization with cycle-consistency constraints further regularizes the manipulated images and preserves details of the source images. Our results achieve better visual quality than previous language-based image editing approaches. 


\section{Related Work}
\noindent\textbf{Image Manipulation with Generative Models.} 
Zhu~\etal~\cite{zhu2016generative} to defines coloring, sketching, and warping brush as editing operations and used constrained optimization to update images. 
%
%
Similarly, Andrew~\etal~\cite{brock2016neural} proposes Introspective Adversarial Network (IAN) which optimizes the latent space to generate manipulated images according to the input images and user brush inputs. 
%
%
%
GANPaint~\cite{bau2019semantic} manipulates the latent space of the input image guided by GAN Dissection~\cite{bau2018gan}, which relies on a segmentation model to identify latent units related to specific objects.
This approach therefore is mainly suitable for adding or removing specific types of objects from images. 
Another line of work focuses on face or fashion attribute manipulation with predefined attributes and labeled images on face or fashion datasets~\cite{perarnau2016invertible,shu2017neural,xiao2018elegant,shen2019interpreting,ak2019attribute,chen2019semantic,usman2019puppetgan}.
In contrast, our approach aims to handle open-vocabulary image manipulation on arbitrary colors, textures, and high-level attributes without attribute annotations for training.
%

\noindent\textbf{Language-based Image Editing.}
%
%
The interaction between language and vision has been studied for various applications~\cite{vinyals2015show,liu2018show,chen2018improving,liu2019improving,wang2019camp}.
Language-based image editing enables user-friendly control for image editing by free-form sentences or phrases as the manipulation instructions. 
~\cite{el2018keep,cheng2018sequential,chen2018language} collects paired data (\ie original images, manipulation queries, and images after manipulation) for training.
However, collecting such data is time-consuming and infeasible for most editing scenarios.
Other works~\cite{zhu2017your,dong2017semantic,nam2018text,gunel2018language,mao2019bilinear,yu2019multi} only require image-caption pairs for training, but those methods are restricted to specific image domains with fine-grained descriptions such as flowers or birds. 
%
%
%
%
Our work extends the problem setting to open-domain images.
Moreover, our approach does not rely on fine-grained accurate captions. Instead, we use Conceptual Captions dataset, where the images and captions are harvested from the web.
Concurrent work~\cite{li2020manigan} conducts language-based image editing on COCO dataset. But it takes a trade-off between reconstructing and editing, restricting the model from achieving both high-quality images and effective editing at the same time.
%
%

\noindent\textbf{Image-to-image Translation.}
Supervised image-to-image translation~\cite{isola2017image,chen2017photographic,wang2018high,park2019semantic,liu2019learning} translates images between different domains with paired training data.
\cite{taigman2016unsupervised,liu2016coupled,royer2017xgan,yi2017dualgan,zhu2017unpaired,kim2017learning} focus on unsupervised translation with unpaired training data. 
Consequent works focus on multi-domain~\cite{choi2018stargan} or multi-modal~\cite{zhu2017toward,almahairi2018augmented,huang2018multimodal,lee2018diverse}. 
%
%
However, one have to define domains and collect domain-specific images for image-to-image translation, which is not able to tackle arbitrary manipulation instructions.
On the contrary, our approach performs open-vocabulary image manipulation without defining domains and collecting domain-specific images.

\section{Method}
Our goal of open-vocabulary open-domain image manipulation is to edit an arbitrary image based on an open-vocabulary manipulation instruction.
The manipulation instructions should indicate the source objects or attributes to be edited as well as the target objects or attributes to be added.
For example, the manipulation instruction could be ``red apple $\rightarrow$ green apple''.

There are several challenges for open-vocabulary open-domain image manipulation: 
(1) It is difficult to obtain a plausible manipulation instruction for each training image. And it is infeasible to collect large-scale ground-truth manipulated images for fully supervised training.
(2) The open-domain images are of high variations, compared with previous work which only consider single-domain images like flowers or birds.
(3) The manipulated images may fail to preserve all details of the source images.
Previous work on language-guided image editing uses other images' captions as the manipulation instruction for an image to train the model.
However, it assumes that all images are from the same domain, while cannot handle open-domain images, \eg, a caption for a flower image cannot be used as the manipulation instruction for a bird image.

\begin{figure*}[t]
\centering
\includegraphics[width=1\linewidth]{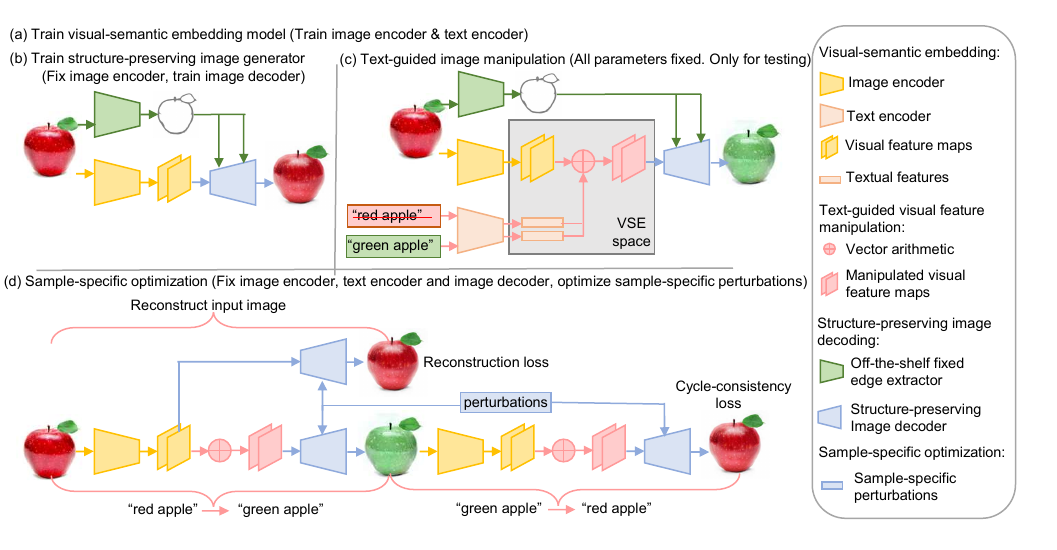}
\caption{The pipeline of our Open-Edit framework. (a) and (b) show the training process. (c) and (d) illustrate the testing process. To simplify the demonstration, edge extractor and text encoder are omitted in (d).}
\label{fig:pipeline}
\end{figure*}

To achieve open-vocabulary open-domain image manipulation, we propose a simple but effective pipeline, named Open-Edit, as shown in Fig.~\ref{fig:pipeline}.
It exploits the visual-semantic joint embedding space to manipulate visual features by textual features, and then decodes the images from the manipulated feature maps.
It is composed of visual-semantic embedding, text-guided visual feature manipulation, structure-preserving image decoding, and sample-specific optimization.

There are two stages for training. In the first stage, we pretrain the \textit{visual-semantic embedding} (VSE) model on a large-scale image-caption dataset to embed any images and texts into latent codes in the visual-semantic embedding space (Fig.~\ref{fig:pipeline}(a)). 
Once trained, the VSE model is fixed to provide image and text embeddings.
In the second stage, the \textit{structure-preserving image decoder} is trained to reconstruct the images from the visual feature maps encoded by the VSE model, as shown in Fig.~\ref{fig:pipeline}(b).
The whole training process only requires the images and noisy captions harvested from the web, and does not need any human annotated manipulation instructions or ground-truth manipulated images.

During inference (Fig.~\ref{fig:pipeline}(c)), the visual-semantic embedding model encodes the input images and manipulation instructions into visual feature maps and textual features in the joint embedding space. 
Then \textit{text-guided visual feature manipulation} is performed to ground the manipulation instructions on the visual feature maps and manipulate the corresponding regions of the visual feature maps with the provided textual features.
Next, the structure-preserving image decoder generates the manipulated images from the manipulated feature maps.

%

%
%
%

Furthermore, in order to regularize the manipulated images and preserve details of the source images, we introduce small sample-specific perturbations added to the intermediate layers of the image decoder, and propose a \textit{sample-specific optimization} approach to optimize the perturbations based on the input image and instruction, shown in Fig.~\ref{fig:pipeline}(d).
For a specific image and manipulation instruction, we put constraint on both the reconstructed images and the images generated by a pair of cycle manipulations (\eg, red apple $\rightarrow$ green apple $\rightarrow$ red apple).
In this way, we adapt the image generator to the specific input image and instruction and achieve higher quality image manipulations.

\subsection{A Revisit of Visual-Semantic Embedding}

To handle open-vocabulary instructions and open-domain images, we use a large-scale image-caption dataset to learn a universal visual-semantic embedding space.
Convolutional neural networks (CNN) and long short-term memory networks (LSTM) are used as encoders to transform images and captions into visual and textual feature vectors.
A triplet ranking loss with hardest negatives, as shown below, is applied to train the visual and textual encoders~\cite{faghri2017vse++}.
%
\begin{equation}\label{eq:vseloss}
\mathcal{L}(\mathbf{v},\mathbf{t})=\max_{\hat{\mathbf t}}[m+\langle\mathbf{v},\hat{\mathbf t}\rangle-\langle\mathbf{v},\mathbf{t}\rangle]_+ + \max_{\hat{\mathbf v}}[m+\langle\hat{\mathbf v}, \mathbf{t}\rangle - \langle\mathbf{v}, \mathbf{t}\rangle]_+
\end{equation}
where $\mathbf{v}$ and $\mathbf{t}$ denote the visual and textual feature vectors of a positive image-caption pair.
$\hat{\mathbf v}$ and $\hat{\mathbf t}$ are the negative image and caption features in the mini-batch.
$[x]_+=\max(0,x)$, and $m$ is the constant margin for the ranking loss.
$\langle\mathbf{v},\mathbf{t}\rangle$ denotes the dot product to measure the similarity between the visual and textual features.
With the trained VSE model, the visual feature maps before average pooling $\mathbf{V} \in \mathbb{R}^{1024\times7\times7}$ is also embedded into the VSE space.

\subsection{Text-guided Visual Feature Manipulation}
%
%
The universal visual-semantic embedding space enables us to manipulate the visual feature maps with the text instructions by vector arithmetic operations, similar to that of word embeddings (\eg, ``king'' - ``man'' + ``woman'' = ``queen'')~\cite{mikolov2013distributed}.
When manipulating certain objects or attributes, we would like to only modify specific regions while keeping other regions unchanged.
So instead of editing the global visual feature vector, we conduct vector arithmetic operations between the visual feature maps $\mathbf{V} \in \mathbb{R}^{1024\times7\times7}$ and textual feature vectors. 
%

\begin{figure}[t]
\centering
\includegraphics[width=1\linewidth]{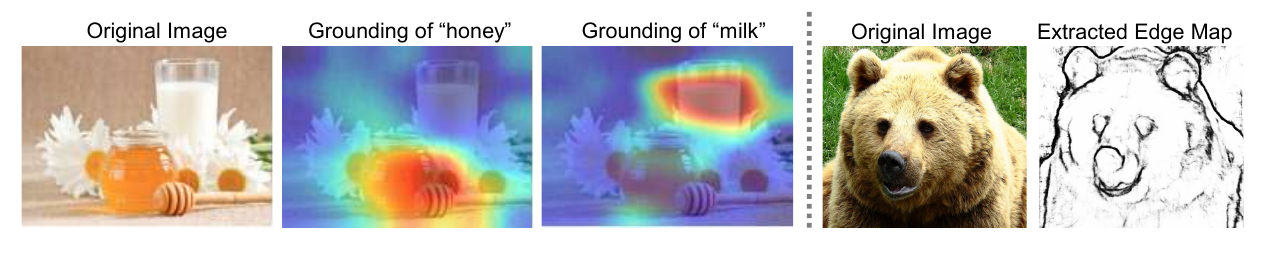}
\caption{Left: an example of grounding results by visual-semantic embedding. Right: an example of edge map extracted by off-the-shelf edge detector.}
\label{fig:grounding}
\end{figure}

We first identify which regions in the feature map to manipulate, \ie, ground the manipulation instructions on the spatial feature map.
The VSE model provides us a soft grounding for textual queries by a weighted summation of the image feature maps, similar to class activation maps (CAM)~\cite{zhou2016learning}.
%
%
We use the textual feature vector $\mathbf{t}\in \mathbb{R}^{1024\times1}$ as weights to compute the weighted summation of the image feature maps $\mathbf{g} = \mathbf{t}^{\top}\mathbf{V}$.
This scheme gives us a soft grounding map $\mathbf{g}\in \mathbb{R}^{7\times7}$, which is able to roughly localize corresponding regions in the visual feature maps related to the textual instruction.
Examples of the text-guided soft grounding results are shown in Fig.~\ref{fig:grounding} (left).
We adopt the grounding map as location-adaptive coefficients to control the manipulation strength at different locations.
We further adopt a coefficient $\alpha$ to control the global manipulation strength, which enables continuous transitions between source images and the manipulated ones.
%
%
The visual feature vector at spatial location $(i,j)$ (where $i,j\in \{0,1,...6\}$) in the visual feature map $\mathbf{V} \in \mathbb{R}^{1024\times7\times7}$, is denoted as $\mathbf{v}^{i,j} \in \mathbb{R}^{1024}$.
We define the following types of manipulations by vector arithmetics weighted by the soft grounding map and the coefficient $\alpha$.

\noindent\textbf{Changing Attributes.}
Changing object attributes or global attributes is one of the most common manipulations.
The textual feature embeddings of the source and target concepts are denoted as $\mathbf{t}_1$ and $\mathbf{t}_2$. respectively.
For example, if we want to change a ``red apple'' into a ``green apple'', $\mathbf{t}_1$ would be the textual embedding of phrase ``red apple'' and $\mathbf{t}_2$ would be the embedding of phrase ``green apple''.
The manipulation of image feature vector $\mathbf{v}^{i,j}$ at location $(i,j)$ is,
\begin{equation}
\mathbf{v}^{i,j}_m=\mathbf{v}^{i,j}-\alpha\langle\mathbf{v}^{i,j},\mathbf{t}_1\rangle\mathbf{t}_1 + \alpha\langle\mathbf{v}^{i,j},\mathbf{t}_1\rangle\mathbf{t}_2,
\end{equation}
where $i,j\in \{0,1,...6\}$, and $\mathbf{v}^{i,j}_m$ is the manipulated visual feature vector at location $(i,j)$ of the $7\times7$ feature map.
We remove the source features $\mathbf{t}_1$ and add the target features $\mathbf{t}_2$ to each visual feature vector $\mathbf{v}^{i,j}$.
$\langle \mathbf{v}^{i,j}, \mathbf{t}_1\rangle$ is the value of the soft grounding map at location $(i,j)$, calculated as the dot product of the image feature vector and the source textual features.
We can also interpret the dot product as the projection of the visual embedding $\mathbf{v}^{i,j}$ onto the direction of the textual embedding $\mathbf{t}_1$.
It serves as a location-adaptive manipulation strength to control which regions in the image should be edited. 
$\alpha$ is a hyper-parameter that controls the image-level manipulation strength.
By smoothly increasing $\alpha$, we can achieve smooth transitions from source to target attributes.

\noindent\textbf{Removing Concepts.}
In certain scenarios, objects, stuff or attributes need to be removed, \eg, remove the beard from a face.
Denote the semantic embedding of the concept we would like to remove as $\mathbf{t}$.
The removing operation is
%
\begin{equation}
\mathbf{v}^{i,j}_m=\mathbf{v}^{i,j}-\alpha\langle\mathbf{v}^{i,j},\mathbf{t}\rangle\mathbf{t}.
\end{equation}

\noindent\textbf{Relative Attributes.}
Our framework also handles relative attribute manipulation, such as making a red apple less red or tuning the image to be brighter.
Denote the semantic embedding of the relative attribute as $\mathbf{t}$.
The strength of the relative attribute is controlled by the hyper-parameter $\alpha$.
By smoothly adjusting $\alpha$, we can gradually strengthen or weaken the relative attribute as
\begin{align}
\mathbf{v}^{i,j}_m&=\mathbf{v}^{i,j}\pm\alpha\langle\mathbf{v}^{i,j},\mathbf{t}\rangle\mathbf{t}.
\end{align}

\subsection{Structure-Preserving Image Decoding}

After deriving the manipulated feature map $\mathbf{V}_m \in \mathbb{R}^{1024\times7\times7}$, an image decoder takes $\mathbf{V}_m$ as input and generates the manipulated images.

Since we do not have paired data for training and it is difficult to generate plausible manipulation instructions for each image, we train the image decoder with only the reconstruction supervisions, as shown in Fig.~\ref{fig:pipeline}(b).
Specifically, we fix the VSE model to transform an image $\mathbf{I}$ into the feature maps $\mathbf{V}$ in the joint embedding space, and train a generative adversarial network to reconstruct the input image from $\mathbf{V}$.
The generator is trained with the hinge-based adversarial loss, discriminator feature matching loss, and perceptual loss.
\begin{align}
\mathcal{L}_G =& - \mathbb{E}[D(G(\mathbf{V}))] + \lambda_{VGG} \mathbb{E}\large[\sum_{k=1}^N\frac{1}{n_k}||F_k(G(\mathbf{V}))-F_k(\mathbf{I})||_1\large]\nonumber\\
            &+ \lambda_{FM}\mathbb{E}\large[\sum_{k=1}^N\frac{1}{m_k}||D_k(G(\mathbf{V}))-D_k(\mathbf{I})||_1\large],\nonumber\\
\mathcal{L}_D &= - \mathbb{E}[\min(0,-1+D(\mathbf{I}))]-\mathbb{E}[min(0,-1-D(G(\mathbf{V})))],
\end{align}
where $\mathcal{L}_D$ and the first term of $\mathcal{L}_G$ are the hinge-based adversarial loss.
The second term of $\mathcal{L}_{G}$ is the perceptual loss, calculated as the VGG feature distance between the reconstructed image and the input image.
The third term of $\mathcal{L}_{G}$ is the discriminator feature matching loss, which matches the intermediate features of the discriminator between the reconstructed image and the input image.
$n_k$ and $m_k$ are the number of elements in the $k$-th layer of the VGG network and discriminator, respectively.
$\lambda_{VGG}$ and $\lambda_{FM}$ are the loss weights.
Although not being trained on manipulated feature maps, the image decoder learns a general image prior. 
So during inference, the decoder is able to generate manipulated images when given the manipulated feature maps as input.

Furthermore, we incorporate edge constraints into the image decoder to preserve the structure information when editing image appearances. 
We adopt an off-the-shelf CNN edge detector~\cite{he2019bi} to extract edges from the input images.
The extracted edges, as shown in Fig.~\ref{fig:grounding} (right), are fed into intermediate layers of the image decoder by spatially-adaptive normalization~\cite{park2019semantic}.
Specifically, we use the edge maps to predict the spatially-adaptive scale and bias parameters of batch-normalization layers.
We denote the edge map as $\mathcal{E}$.
Denote the feature map value of the $n$-th image in the mini-batch at channel $c$ and location $(h,w)$ as $f_{n,c,h,w}$.
Denote the mean and standard deviation of the feature maps at channel $c$ as $\mu_c$ and $\sigma_c$, respectively.
The spatially-adaptive normalization is
%
\begin{equation}
\gamma_{c,h,w}(\mathcal{E})\frac{f_{n,c,h,w}-\mu_c}{\sigma_c}+\beta_{c,h,w}(\mathcal{E}),
\end{equation}
where $\gamma$ and $\beta$ are two-layer convolutions to predict spatially-adaptive scale and bias for BN layers.
With the edge constraints, the decoder is able to preserve the structures and edges of the source images when editing the image appearances.
%

\subsection{Sample-Specific Optimization with Cycle-Consistency Constraints}

The vector arithmetic manipulation operations may not be precise enough, because some attributes might be entangled and the visual-semantic embedding space may not be strictly linear.
%
%
Moreover, the image decoder trained with only reconstruction supervision is not perfect and might not be able to reconstruct all details of the source image.
To mitigate those problems, we adopt a sample-specific optimization approach to adapt the decoder to the specific input image and manipulation instruction.

%
For each image and manipulation instruction (\eg, ``red apple'' $\rightarrow$ ``green apple''), we apply a pair of cycle manipulations to exchange the attributes forth and back (\eg, ``red apple'' $\rightarrow$ ``green apple'' $\rightarrow$ ``red apple'').
The corresponding source and manipulated images are denoted as $\mathbf{I}\rightarrow\mathbf{I}_m\rightarrow\mathbf{I}_c$.
We incorporate a cycle-consistency loss to optimize the decoder to adapt to the specific image and manipulation instruction.
In this way, we can regularize the manipulated image and complete the details missed during encoding and generating.
We also adopt a reconstruction loss to force the optimized decoder to reconstruct the source image without manipulating the latent visual features.
The reconstruction loss $\mathcal{L}_{rec}$ and cycle-consistency loss $\mathcal{L}_{cyc}$ are the summation of $L_1$ loss and perceptual loss, computed between the source image $\mathbf{I}$ and the reconstructed $\mathbf{I}_{r}$ or the cycle manipulated image $\mathbf{I}_{c}$,
%
\begin{align}
\mathcal{L}_{cyc}&=||\mathbf{I}_{c}-\mathbf{I}||_1+\lambda\sum_{k=1}^N\frac{1}{n_k}||F_k(\mathbf{I}_{c})-F_k(\mathbf{I})||_1,\\
\mathcal{L}_{rec}&=||\mathbf{I}_{r}-\mathbf{I}||_1+\lambda\sum_{k=1}^N\frac{1}{n_k}||F_k(\mathbf{I}_{r})-F_k(\mathbf{I})||_1,
\end{align}
where $\lambda$ is the loss weight for perceptual loss and $F_k$ is the $k$-th layer of the VGG network with ${n_k}$ features.

However, directly finetuning the decoder parameters for a specific image and manipulation instruction would cause severe overfitting to the source image, and the finetuned decoder would not be able to generate the high-quality manipulated image.
Alternatively, we fix the decoder parameters and only optimize a series of additive perturbations of the decoder network, as shown in Fig.~\ref{fig:pipeline}(d).
For each specific image and manipulation, the sample-specific perturbations are initialized as zeros and added to the intermediate layers of the decoder.
The perturbation parameters are optimized with the manipulation cycle-consistency loss and reconstruction loss on that specific image and manipulation instruction.
So when generating the manipulated images, the optimized perturbations can complete the high-frequency details of the source images, and regularize the manipulated images.
Specifically, the image decoder is divided into several decoder blocks $G_1,G_2,\cdots,G_n$ ($n=4$ in our implementation), and the perturbations are added to the decoder between the $n$ blocks,
\begin{equation}
G^\prime(\mathbf{V})=G_n(G_{n-1}(\cdots (G_1(\mathbf{V})+\mathbf{P_1})\cdots)+\mathbf{P_{n-1}}),
\end{equation}
where $\bf P_1, \cdots, \bf P_{n-1}$ are the introduced perturbations.
We optimize the perturbations by the summation of reconstructions loss, manipulation cycle-consistency loss, and a regularization loss $\mathcal{L}_{reg}=\sum_{i=1}^{n-1}||\mathbf{P}_i||_2^2$.

Those optimization steps are conducted only during testing. We adapt the perturbations to the input image and manipulation instruction by the introduced optimization process.
Therefore, the learned sample-specific perturbations models high-frequency details of the source images, and regularizes the manipulated images.
In this way, the generator with optimized perturbations is able to generate photo-realistic and detail-preserving manipulated images.

\section{Experiments}

\subsection{Datasets and Implementation Details}

Our visual-semantic embedding model (including image encoder and text encoder) and image decoder are trained on Conceptual Captions dataset~\cite{sharma2018conceptual} with 3 million image-caption pairs harvested from the web.
The images are from various domains and of various styles, including portrait, objects, scenes, and others.
Instead of human-annotated fine-grained descriptions in other image captioning datasets, the captions of Conceptual Captions dataset are harvested from the Alt-text HTML attribute associated with web images.
Although the images are of high variations and the captions are noisy, results show that with large datasets, the VSE model is able to learn an effective visual-semantic embedding space for image manipulation. The image decoder trained with images from Conceptual Captions dataset learns a general image prior for open-domain images.
The model structure and training process of the VSE model follow that of VSE++~\cite{faghri2017vse++}.
The image decoder takes $1024\times7\times7$ feature maps as input, and is composed of 7 ResNet Blocks with upsampling layers in between, which generates $256\times256$ images.
The discriminator is a Multi-scale Patch-based discriminator following~\cite{park2019semantic}.
The decoder is trained with GAN loss, perceptual loss, and discriminator feature matching loss.
The edge extractor is an off-the-shelf bi-directional cascade network~\cite{he2019bi} trained on BSDS500 dataset~\cite{arbelaez2010contour}.

\subsection{Applications and Results}
\begin{figure*}[t]
\centering
\includegraphics[width=1\linewidth]{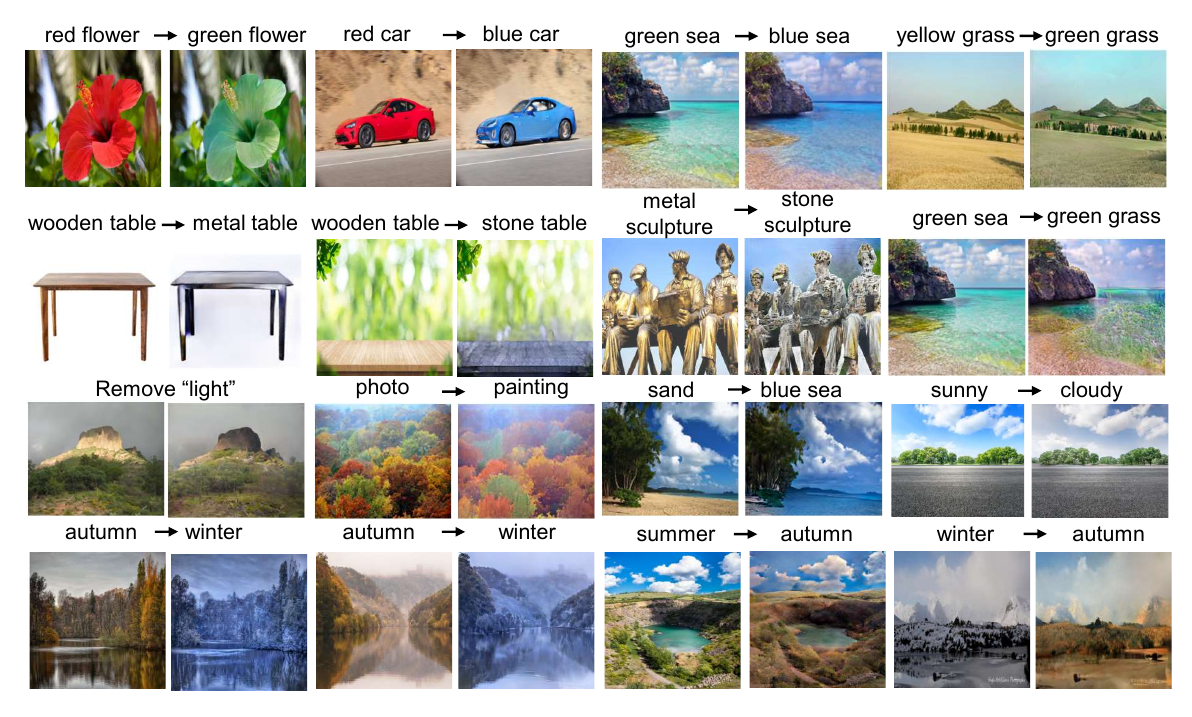}
\caption{Applications and manipulation results of our method.}
\label{fig:result}
\end{figure*}

Our approach can achieve open-domain image manipulation with open-vocabulary instructions, which has various applications.
We demonstrate several examples in Fig.~\ref{fig:result}, including changing color, texture, and global or local high-level attributes.

Results in the first row demonstrate that our model is able to change \textit{color} for objects and stuff while preserving other details of the image.
Moreover, it preserves the lighting conditions and relative color strengths very well when changing colors.
Our model is also able to change \textit{textures} of the images with language instructions, for example, editing object materials or changing sea to grass, as shown in the second row.
Results indicate that the VSE model learns effective texture features in the joint embedding space, and that the generator is able to generate reasonable textures based on the manipulated features.
Besides low-level attributes, our model is also able to handle \textit{high-level semantic attributes}, such as removing lights, changing photos to paintings, sunny to cloudy, and transferring seasons, in the third and fourth rows.

\begin{figure}[t]
\centering
\includegraphics[width=1\linewidth]{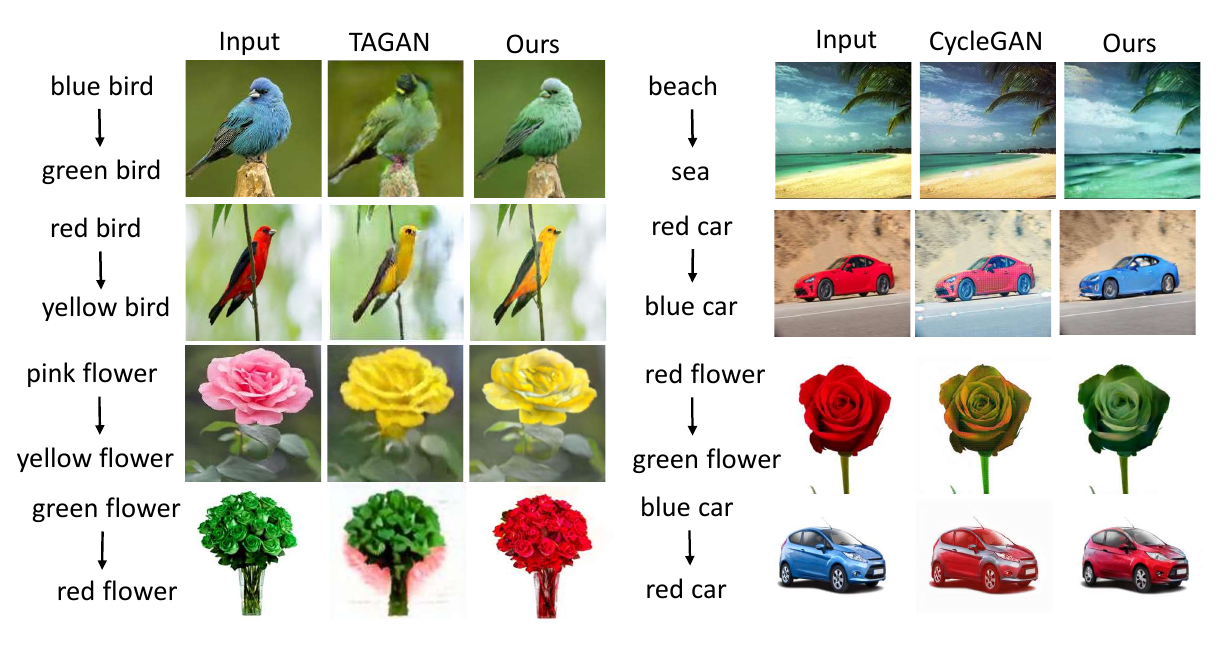}
\caption{Comparison with previous language-based editing method TAGAN~\cite{nam2018text} (left) and image-to-image translation method CycleGAN~\cite{zhu2017unpaired} (right).}
\label{fig:compare}
\end{figure}

\noindent\textbf{Quantitative evaluation.}
Since ground-truth manipulated images are not available, we conduct evaluations by user study, L2 error, and LPIPS.

The user study is conducted to evaluate human perceptions of our approach.
For each experiment, we randomly pick 60 images and manually choose the appropriate manipulation instructions for them.
The images cover a wide variety of styles and the instructions range from color and texture manipulation to high-level attribute manipulation.
10 users are asked to score the 60 images for each experiment by three criteria, (1) visual quality, (2) how well the manipulated images preserve the details of the source image, and (3) how well the manipulation results match the instruction.
The scores range from 1 (worst) to 5 (best), and we will analyze the results shown in Table.~\ref{tab:user_study} in the following.

To evaluate the visual quality and content preservation, we calculate the L2 error and Perceptual similarity (LPIPS)~\cite{zhang2018unreasonable} between the reconstructed images and input images, as shown in Table~\ref{tab:l2error} and analyzed in the following.

\noindent\textbf{Comparison with previous work.}
Since this is the first work to explore open-domain image manipulation with open-vocabulary instructions, our problem setting is much more challenging than previous approaches.
Nonetheless, we compare with two representative approaches, CycleGAN~\cite{zhu2017unpaired} and TAGAN~\cite{nam2018text}.

CycleGAN is designed for image-to-image translation, but we have to define domains and collect domain-specific images for training. So it is not able to tackle open-vocabulary instructions.
To compare with CycleGAN, we train three CycleGAN models for translating between blue and red objects, translating between red and green objects, and translating between beach and sea, respectively.
The images for training CycleGAN are retrieved from Conceptual Captions with our visual-semantic embedding model.
Qualitative comparison are shown in Fig.~\ref{fig:compare}, and user study are shown in Table~\ref{tab:user_study}.
Results indicate that both our approach and CycleGAN is able to preserve details of the input images very well, but CycleGAN worse at transferring desired attributes in some scenarios.

State-of-the-art language-based image manipulation method TAGAN~\cite{nam2018text} uses mismatched image-caption pairs as training samples for manipulation.
It is able to handle language instructions, but is limited to only one specific domain such as flowers (Oxford-102) or birds (CUB).
It also requires fine-grained human-annotated descriptions of each image in the dataset.
While our approach handles open-domain images with noisy captions harvested from the web for training.
Since TAGAN only has the models for manipulating bird or flower images, we compare our results with theirs on flower and bird image manipulation in Fig.~\ref{fig:compare}.
We also compare user evaluation in Table~\ref{tab:user_study}. Quantitative evaluation of L2 error and perceptual metric (LPIPS) between reconstructed images and original images are shown in Table~\ref{tab:l2error}.
Our model is not trained on the Oxford-102 or CUB datasets, but still performs better than the TAGAN models specifically trained on those datasets.
Moreover, the L2 reconstruction error also shows that our model has the potential to preserve the detailed contents of the source images.

\begin{table}[t]
\caption{User study results on visual quality, how well the manipulated images preserve details, and how well the manipulated images match the instruction. In the table, ``edge'' represents edge constraints in the image decoder, and ``opt'' represents the sample-specific optimization.}\label{tab:user_study}
\scriptsize
\begin{tabular}{c|c|c|c|c|c}
\hline
                           & CycleGAN\cite{zhu2017unpaired} & TAGAN\cite{nam2018text} & w/o edge, w/o opt & w/ edge, w/o opt & \textbf{w/ edge, w/ opt} \\ \hline
Visual quality     & 4.0    & 3.1       & 1.3           & 4.1             & \textbf{4.4}              \\
Preserve details   & 4.2    & 2.7       & 1.2           & 3.7             & \textbf{4.3}              \\
Match instruction  & 1.9    & 4.2       & 4.0           & 4.5             & \textbf{4.5}     \\ \hline
\end{tabular}
\footnotesize
\caption{L2 error and LPIPS between reconstructed and original images of TAGAN and ablations of our approach. Lower L2 reconstruction error and LPIPS metric indicates that the reconstructed images preserve details of the source images better.}\label{tab:l2error}
\scriptsize
\begin{tabular}{c|c|c|c|c}
\hline
                                & TAGAN\cite{nam2018text} & w/o edge, w/o opt & w/ edge, w/o opt & \textbf{w/ edge, w/ opt} \\ \hline
L2 error on Oxford-102 test set  			& 0.11       			  & 0.19              & 0.10             & \textbf{0.05}              \\
L2 on Conceptual Captions val     & N/A       			  & 0.20           	  & 0.12             & \textbf{0.07}              \\ 
LPIPS on Conceptual Captions val        & N/A                     & 0.33              & 0.17             & \textbf{0.06}             \\ \hline
\end{tabular}
\end{table}

\begin{figure}[t]
\centering
\includegraphics[width=1\linewidth]{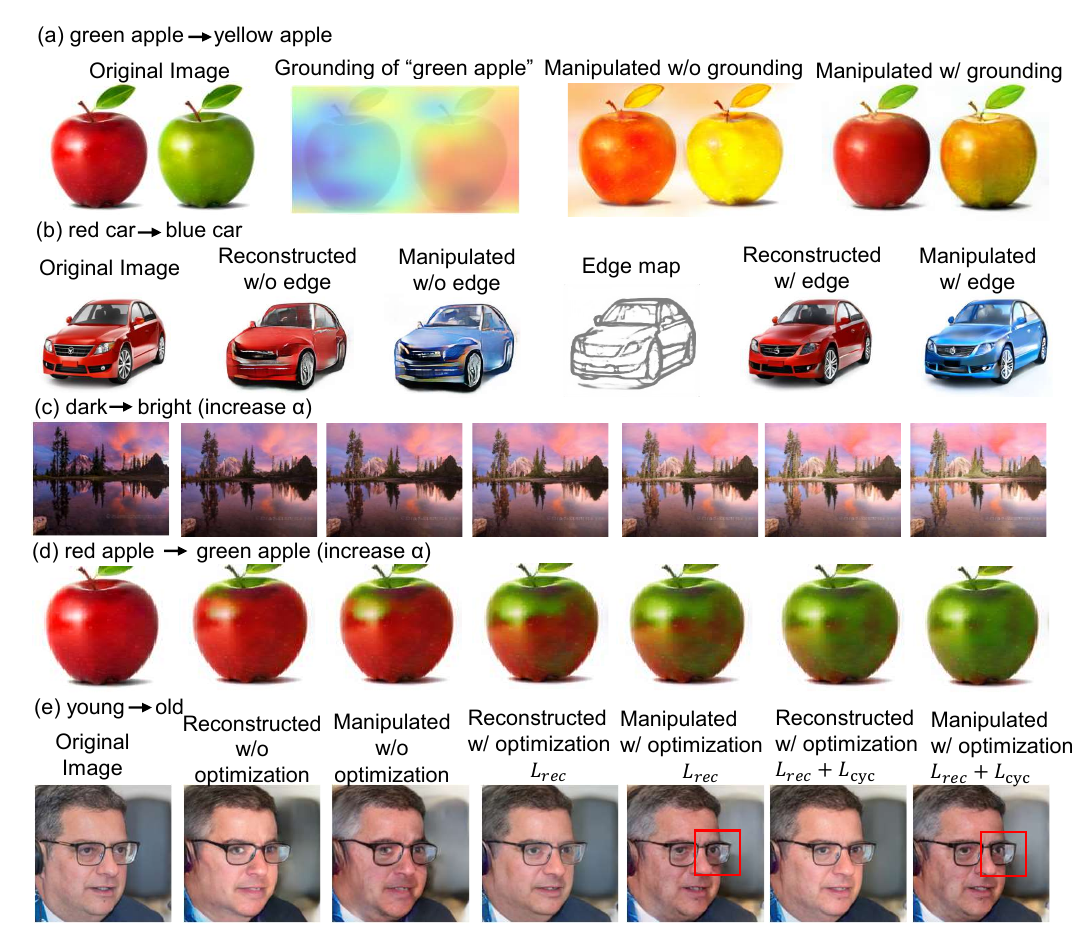}
\caption{Component analysis of our approach. The examples from top to bottom show analysis on grounding, edge constraints, adjusting coefficient, and the sample-specific optimization and cycle-consistency constraints, respectively.}
\label{fig:ablation}
\end{figure}

\subsection{Component Analysis}

\noindent\textbf{The effectiveness of instruction grounding.}
Our text-guided visual feature manipulation module uses the soft instruction grounding maps as location-adaptive manipulation coefficients to control the local manipulation strength at different locations.
The instruction grounding map is very important when the manipulation instruction is related to local areas or objects in the image.
Fig.~\ref{fig:ablation}(a) demonstrates the effectiveness of adopting grounding for manipulation, where we aim to change the green apple into a yellow apple and keep the red apple unchanged.
The grounding map is able to roughly locate the green apple, and with the help of the soft grounding map, the model is able to change the color of the green apple while keeping the red apple and the background unchanged.
On the contrary, the model without grounding changes not only the green apple, but also the red apple and the background.

\noindent\textbf{Edge Constraints.}
Our structure-aware image decoder exploits edge constraints to preserve the structure information when generating the reconstructed and manipulated images.
In Fig.~\ref{fig:ablation}(b), we show an example of image reconstruction and manipulation with and without edges.
The image decoder is able to reconstruct and generate higher-quality images with clear structures and edges with edge constraints.
User study results in Table~\ref{tab:user_study} and quantitative evaluation in Table~\ref{tab:l2error} also indicate that the generated images are of better visual quality and preserve details better with the structure-aware image decoder.

\noindent\textbf{Adjusting coefficient $\alpha$ for smooth attribute transition.}
The hyper-parameter $\alpha$ controls the global attribute manipulation strength, which can be adjusted according to user requirements.
By gradually increasing $\alpha$, we obtain a smooth transition from the source images to the manipulated images with different manipulation strengths.
Fig.~\ref{fig:ablation}(c)(d) illustrates the smooth transition of an image from dark to bright, and from red apple to green apple, respectively.

\noindent\textbf{The effectiveness of sample-specific optimization and cycle-consistency constraints.}
Fig.~\ref{fig:ablation}(e)\footnote{The decoder for Fig.~\ref{fig:ablation}(e) is trained on FFHQ~\cite{karras2018style} to learn the face image prior.} demonstrates the effectiveness of sample-specific optimization and cycle-consistency constraints.
The reconstructed image and manipulated image without sample-specific optimization miss some details such as the shape of the glasses.
With the perturbation optimization by reconstruction loss, our model is able to generate better-quality reconstructed and manipulated images.
Optimizing the perturbations with both reconstruction loss and manipulation cycle-consistency loss further improves the quality of the generated images, \eg, the glasses are more realistic and the person identity appearance is better preserved.
User study in Table~\ref{tab:user_study} and quantitative evaluation in Table~\ref{tab:l2error} indicate that the sample-specific optimization has the potential of enhancing details of the generated images.

\section{Conclusion and Discussions}
We propose Open-Edit, the first framework for open-vocabulary open-domain image manipulation with minimal training supervision.
It takes advantage of the pretrained visual-semantic embedding, and manipulates visual features by vector arithmetic with textual embeddings in the joint embedding space.
The sample-specific optimization further regularizes the manipulated images and encourages realistic and detail-preserving results.
Impressive color, texture, and semantic attribute manipulation are shown on various types of images.

We believe that this is a challenging and promising direction towards more general and practical image editing, and that our attempt would inspire future work to enhance editing qualities and extend the application scenario. 
In this work we focus on editing appearance-related attributes without changing the structure of images. Further work can be done on more challenging structure-related editing and image editing with more complex sentence instructions.

\section*{Acknowledgement}
This work is supported in part by the General Research Fund through the Research Grants Council of Hong Kong under Grants CUHK14202217, CUHK14203118, CUHK14205615, CUHK14207814, CUHK14213616, CUHK14207319, CUHK14208619, and in part by Research Impact Fund R5001-18.

%
%
\bibliographystyle{splncs04}
\bibliography{egbib}
\end{document}